\documentclass[sigconf]{acmart}
\AtBeginDocument{%
  \providecommand\BibTeX{{%
    \normalfont B\kern-0.5em{\scshape i\kern-0.25em b}\kern-0.8em\TeX}}}

\setcopyright{acmcopyright}
\copyrightyear{2018}
\acmYear{2018}
\acmDOI{10.1145/1122445.1122456}

\acmConference[Woodstock '18]{Woodstock '18: ACM Symposium on Neural
  Gaze Detection}{June 03--05, 2018}{Woodstock, NY}
\acmBooktitle{Woodstock '18: ACM Symposium on Neural Gaze Detection,
  June 03--05, 2018, Woodstock, NY}
\acmPrice{15.00}
\acmISBN{978-1-4503-XXXX-X/18/06}

\usepackage{graphicx}
\usepackage{latexsym,amsmath}
\usepackage{subfigure}
\usepackage{xspace}
\usepackage{multirow}
\usepackage{algorithm}
\usepackage{algorithmic}

\newcommand{\model}{CP-Tuning\xspace}
\newcommand{\fullmodel}{Contrastive Prompt Tuning\xspace}

\newcommand\blfootnote[1]{%
  \begingroup
  \renewcommand\thefootnote{}\footnote{#1}%
  \addtocounter{footnote}{-1}%
  \endgroup
}

\begin{document}

\title{Making Pre-trained Language Models End-to-end Few-shot Learners with \fullmodel}

\author{Ziyun Xu$^{1,2*}$, Chengyu Wang$^{1*}$, Minghui Qiu$^{1\#}$, Fuli Luo$^{1}$, Runxin Xu$^{1,3}$, Songfang Huang$^{1}$, Jun Huang$^{1}$}
\affiliation{%
  \institution{$^{1}$ Alibaba Group $^{2}$ School of Computer Science, Carnegie Mellon University $^{3}$ Key Laboratory of Computational Linguistics, Peking University}
  \country{}}
\email{ziyunx@andrew.cmu.edu,{chengyu.wcy,minghui.qmh,lfl259702,songfang.hsf,huangjun.hj}@alibaba-inc.com,runxinxu@gmail.com}





\begin{abstract}
Pre-trained Language Models (PLMs) have achieved remarkable performance for various language understanding tasks in IR systems, which require the fine-tuning process based on labeled training data. For low-resource scenarios, prompt-based learning for PLMs exploits prompts as task guidance and turns downstream tasks into masked language problems for effective few-shot fine-tuning.
In most existing approaches, the high performance of prompt-based learning heavily relies on handcrafted prompts and verbalizers, which may limit the application of such approaches in real-world scenarios.
To solve this issue, we present~\emph{\model}, the first end-to-end~\emph{\fullmodel} framework for fine-tuning PLMs~\emph{without any manual engineering of task-specific prompts and verbalizers}. It is integrated with the task-invariant continuous prompt encoding technique with fully trainable prompt parameters. 
 We further propose 
the pair-wise cost-sensitive contrastive learning procedure to optimize the model in order to achieve verbalizer-free class mapping and enhance the task-invariance of prompts.
It explicitly learns to distinguish different classes and makes the decision boundary smoother by assigning different costs to easy and hard cases.
Experiments over a variety of language understanding tasks used in IR systems and different PLMs show that~\emph{\model} outperforms state-of-the-art methods.
\blfootnote{$^{*}$~Z. Xu and C. Wang contributed equally. Corresponding author: M. Qiu.}
\end{abstract}




\maketitle

\section{Introduction}

Starting from BERT~\cite{DBLP:conf/naacl/DevlinCLT19}, fine-tuning large-scale Pre-trained Language Models (PLMs) has become the~\emph{de facto} standard practice for solving a majority of Natural Language Processing (NLP) tasks~\cite{DBLP:conf/nips/YangDYCSL19,DBLP:conf/iclr/LanCGGSS20,DBLP:journals/corr/abs-2107-02137}, which have been extensively used in Information Retrieval (IR) systems for tasks such as content analysis, question matching and question answering~\cite{DBLP:conf/cikm/MacdonaldTM21}. To guarantee high accuracy, it is necessary to obtain a sufficient amount of training data for downstream tasks, which is the bottleneck in low-resource scenarios.

The successful application of the ultra-large GPT-3 model~\cite{DBLP:conf/nips/BrownMRSKDNSSAA20} shows that with a sufficiently large memory capacity and massive pre-training computation, large PLMs can learn to solve a task with very few training samples. However, the large model size and the long inference time make it infeasible to deploy such PLMs online with limited computational resources. 
Inspired by these works,~\citet{DBLP:conf/acl/GaoFC20} propose a prompt-based approach to fine-tune BERT-style PLMs in a few-shot learning setting. It converts text classification and regression problems into masked language problems where the knowledge captured during pre-training can be better utilized during the few-shot learning process.
Similar usage of prompts for fine-tuning PLMs has also been shown in~\cite{DBLP:conf/eacl/SchickS21,DBLP:conf/naacl/SchickS21} and many others.~\citet{DBLP:conf/naacl/ScaoR21} conduct a rigorous test to show that prompting is highly beneficial in low-data regimes.

\begin{figure}
\centering
\includegraphics[width=.5\textwidth]{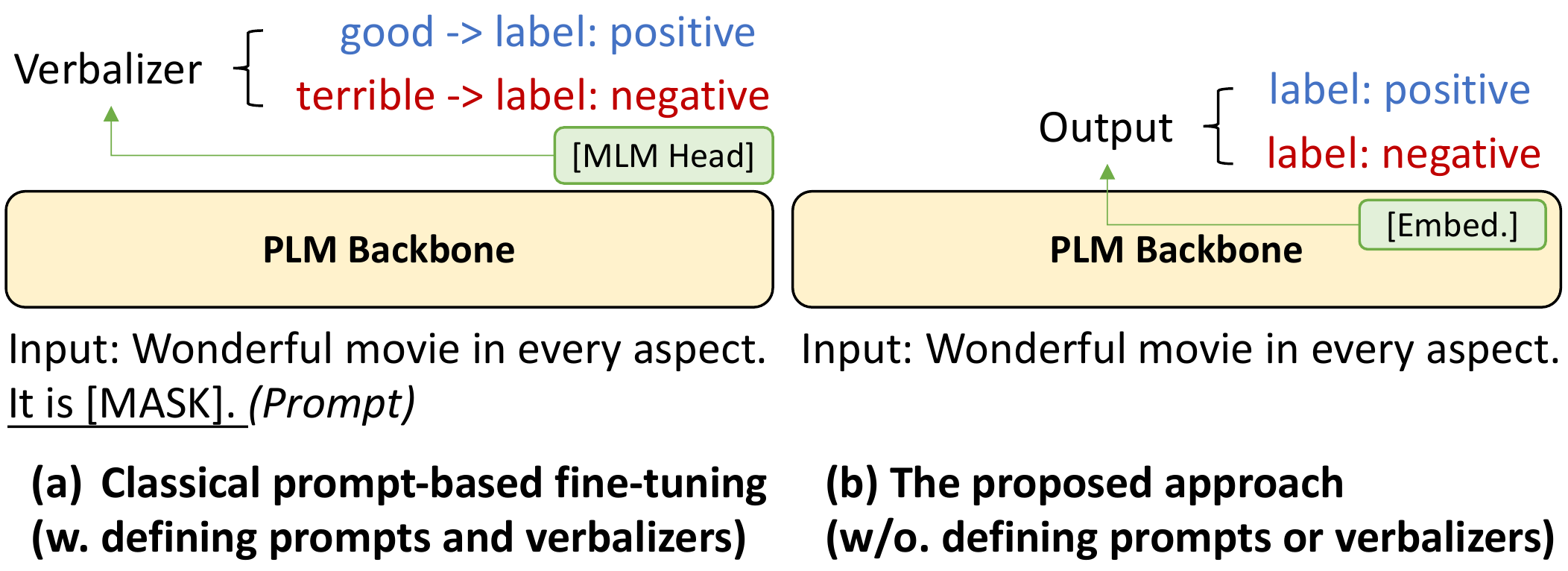}
\vspace{-.5em}
\caption{A simple comparison between classical prompt-based fine-tuning and~\emph{\model} w.r.t. the inputs and outputs. The underlying task is review sentiment analysis.}
\label{fig:intro}
\end{figure}

In most prompt-based approaches, there exist two types of model components that require careful manual engineering, namely~\emph{prompts} and \emph{verbalizers}. Here, prompts are fixed templates or patterns that are employed to inject task-specific guidance to input texts, while verbalizers establish explicit mappings between output tokens and class labels. An example of prompts and verbalizers on review sentiment analysis is illustrated in the left part of Figure~\ref{fig:intro}. As reported in~\cite{DBLP:journals/corr/abs-2103-10385}, designing high-performing prompts and the corresponding verbalizers is challenging and requires a very large validation set. As for prompts, even a slight change of expressions can lead to big variance in the performance of downstream tasks. To alleviate this issue,~\citet{DBLP:journals/corr/abs-2103-10385,DBLP:journals/corr/abs-2110-07602} propose P-tuning, which uses~\emph{continuous prompt embeddings} to avoid the manual prompt engineering process.
However, this method still requires the design of verbalizers, with a strong hypothesis of token-to-label mappings.
Therefore, the drawbacks of prompt and verbalizer engineering potentially hinder the wide application of these approaches. 

To address these issues, we present~\emph{\model}, an end-to-end~\emph{\fullmodel} framework for PLMs~\emph{without the manual design of task-specific prompts and verbalizers}.
To our knowledge, our work is the first to study contrastive learning for prompt-based fine-tuning without manual prompt and verbalizer engineering. Specifically, our approach consists of two major techniques:
\begin{itemize}
\item\textbf{Task-invariant Continuous Prompt Encoding.} We employ~\emph{continuous embeddings} as prompts and do not employ any prompt encoders to avoid learning additional parameters during few-shot learning (in contrast to~\cite{DBLP:journals/corr/abs-2103-10385}).
Specially, we initialize continuous prompt embeddings as the pre-trained representations of a collection of \emph{task-invariant} tokens, and enable prompt embeddings to be~\emph{task-adaptive} by back propagation. Hence,~\emph{\model} does not require manual prompt engineering for specific tasks.

\item\textbf{Verbalizer-free Class Mapping.}
We further propose the novel~\emph{verbalizer-free} mechanism to ease the manual labor of designing verbalizers and to improve the generalization ability of our model, as well as the task-invariance of prompts.
Specifically, the~\emph{Pair-wise Cost-sensitive Contrastive Loss} (\emph{PCCL}) is introduced to train our few-shot learner, together with an auxiliary Mask Language Modeling (MLM) task as the regularizer. 
\emph{PCCL} explicitly learns to distinguish different classes and makes the decision boundary smoother by assigning different costs to easy and hard cases.
In contrast to previous approaches, embeddings of instances before the MLM classifier are directly used for inference. We also theoretically prove that \emph{PCCL} is an extension to various metric learning based loss functions.
\end{itemize}

For evaluation, we conduct extensive experiments to verify the effectiveness of~\emph{\model} over eight public datasets, including various tasks used in IR and NLP systems (e.g., sentiment analysis, question matching, Natural Language Inference (NLI)). Experimental results show that~\emph{\model} consistently outperforms state-of-the-arts for prompt-based few-shot learning.

In summary, we make the following contributions:
\begin{itemize}
    \item We introduce the end-to-end~\emph{\model} framework to enable prompt-based few-shot learning without designing task-specific prompts and verbalizers. To our knowledge, our work is the first to employ contrastive learning for end-to-end prompt-based learning that eases manual engineering.
    \item In~\emph{\model}, the task-invariant continuous prompt encoding technique is presented. We further propose the~\emph{PCCL} technique to train the model without the usage of any verbalizers based on contrastive learning.
    \item Experiments over eight public datasets show that~\emph{\model} consistently outperforms state-of-the-arts for prompt-based few-shot learning. We also theoretically derive the relations between~\emph{PCCL} and other losses.
\end{itemize}

The rest of this paper is organized as follows. Section~\ref{sec:method} presents our~\emph{\model} in detail. The experiments are shown in Section~\ref{sec:exp}, with the related work summarized in Section~\ref{sec:related}. Finally, we draw the conclusion and discuss the future work in Section~\ref{sec:con}.

\section{\emph{\model}: Proposed Approach}

\label{sec:method}

In this section, we begin with an overview of our approach. After that, the detailed techniques are elaborated.

\begin{figure}
\centering
\includegraphics[width=.5\textwidth]{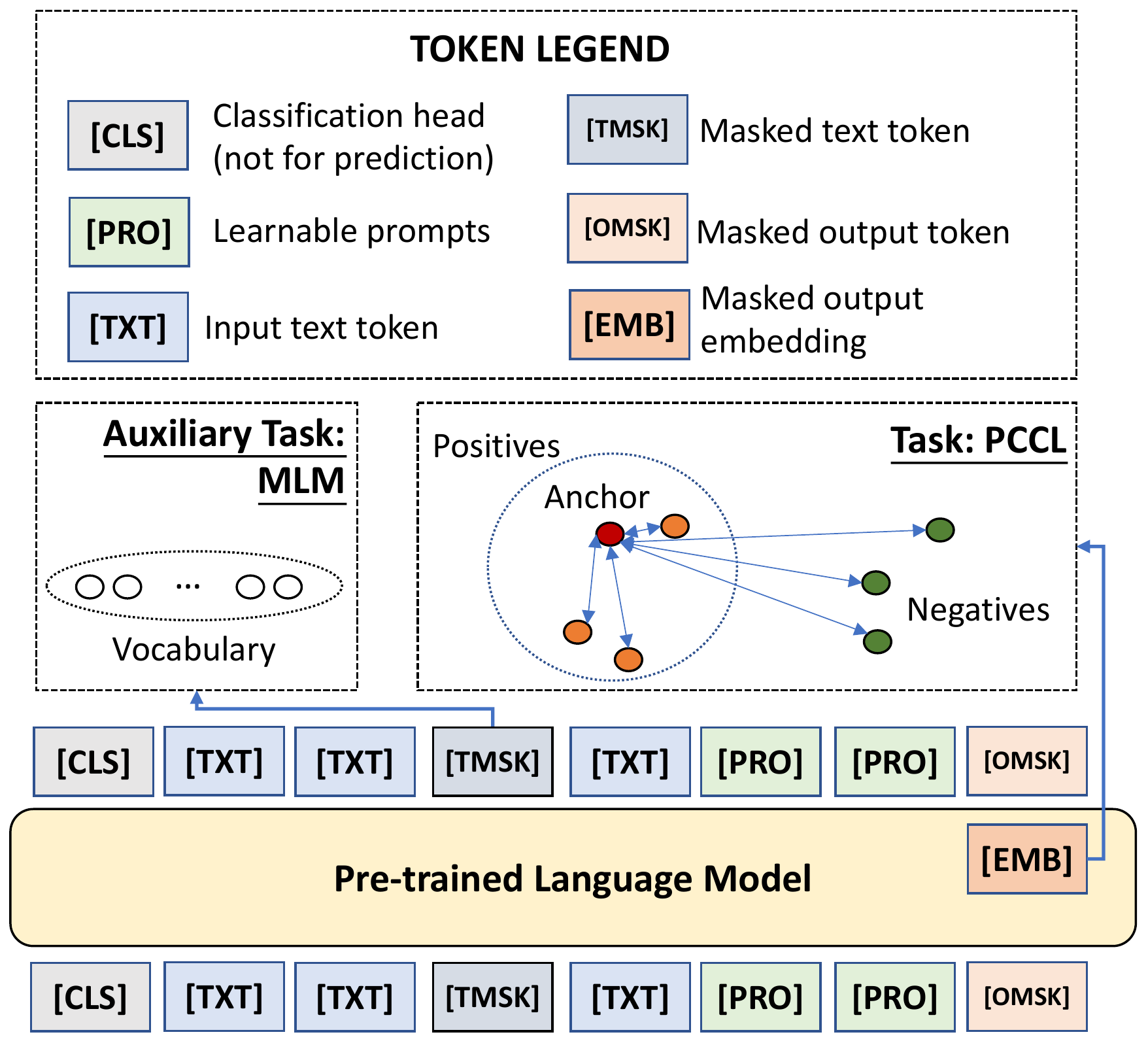}
\vspace{-.5em}
\caption{An overview of the~\emph{\model} framework. For simplicity, we only show input/output text sequences for single-sentence classification tasks.}
\label{fig:framework}
\end{figure}

\begin{table*}
\centering
\begin{small}
\begin{tabular}{l l l}
\hline
\bf Category & \bf Task  & \bf Example of Input Sequence\\
\hline
\hline
Single & Sentiment & Movie fans, get ready to take off... the other direction.\\
Sentence & Analysis & [CLS] Movie [TMSK], get ready to take off... the other direction. [PRO]$_{\text{it}}$ [PRO]$_{\text{is}}$ [OMSK]\\
\hline
Single & Subjectivity & Zero, who, like many Hong Kong youngsters, has a handful of unsteady jobs.\\
Sentence & Classification & [CLS] Zero, who, like [TMSK] Hong Kong youngsters, has a handful of unsteady jobs. [PRO]$_{\text{it}}$ [PRO]$_{\text{is}}$ [OMSK]\\
\hline
\hline
Sentence & NLI & a. What was Telenet? b. Telenet was incorporated in 1973 and started operations in 1975.\\
Pair & & [CLS] What was Telenet? [PRO]$_{\text{?}}$[OMSK][SEP] Telenet was [TMSK] in 1973 and started operations in 1975.\\
\hline
Sentence & Question & a. How do I start trying to trace my family tree? b. How would I start tracing my family history?\\
Pair & Matching & [CLS] How do I start trying to trace my family tree? [PRO]$_{\text{?}}$[OMSK][SEP] How would I [TMSK] tracing my family history?\\
\hline
Sentence & Sentence & a. Around 1,500 police are to be deployed at Niigata for the ferry's visit.	b. About 1,500 police will be deployed for the visit.\\
Pair & Equivalence & [CLS] Around 1,500 police are to be deployed at Niigata for the ferry's [TMSK]. [PRO]$_{\text{?}}$[OMSK][SEP] About 1,500 police\\
& & will be deployed for the visit.\\
\hline
\end{tabular}
\end{small}
\caption{\label{tab:pexample}
Examples of inputs and processed token sequences for~\emph{\model} (first and second lines of input sequence).
[PRO]$_{\text{test}}$ refers to the prompt embedding token initialized by the representation of the token ``text'', which can be updated via back propagation.
Note that the initializations of prompts w.r.t. all single-sentence (or sentence-pair) tasks are the same.
}
\vspace{-.25em}
\end{table*}

\subsection{Overview of~\emph{\model}}

We begin with a few basic notations. 
Let $\mathcal{D}$ be an $N$-way $K$-shot training set of a specific NLP task, where each of the $N$ classes is associated with $K$ training samples.~\footnote{Our work can be easily extended to other fine-tuning scenarios without modification where each class is associated with different numbers of training samples. We also find that~\emph{\model} is better at learning with unbalanced training sets than previous prompt-based methods. Readers can refer to experiments for details.} Denote $\mathcal{M}$ as the collection of parameters of the underlying PLM. The goal of our work is to generate a high-performance few-shot learner initialized from $\mathcal{M}$ based on $\mathcal{D}$ that can effectively generalize to previously unseen data samples of the same task. 
We present the overview of our approach in Figure~\ref{fig:framework}, with major techniques summarized below.

As traditional prompt-based models require the cumbersome process of prompt engineering, we employ~\emph{continuous embeddings} as input prompts. 
Rather than employing sub-networks (\emph{e.g.,}~LSTMs) as prompt encoders~\cite{DBLP:journals/corr/abs-2103-10385}, to avoid learning additional parameters during few-shot learning, we directly feed prompt embeddings to the PLM encoder, and enable the embeddings to be~\emph{task-adaptive} by back propagation.

Besides manually-designed patterns, previous methods also require handcrafted verbalizers, which map the output of the masked token to the class label~\cite{DBLP:conf/eacl/SchickS21,DBLP:conf/naacl/SchickS21}. In our work, we propose the~\emph{verbalizer-free} mechanism to ease the manual labor and to improve the generalization ability of our few-shot learner. 
As prompts and verbalizers are semantically correlated, this technique also  enhances the task-invariance of prompts.
Inspired by the contrastive learning paradigm~\cite{DBLP:journals/corr/abs-2011-0036}, we propose the~\emph{Pair-wise Cost-sensitive Contrastive Loss} (\emph{PCCL}) to train our few-shot learner. In the few-shot learning setting, the lack of training data may easily result in model over-fitting. Hence, an auxiliary MLM loss is also optimized during few-shot learning to alleviate the issue.
In addition, we further show that~\emph{PCCL} is an extension to a variety of loss functions.

\subsection{Task-invariant Prompt Encoding}

The input format of our approach is significantly different from previous works to facilitate~\emph{task-invariant continuous prompt learning}. To be more specific, in contrast to~\cite{DBLP:conf/naacl/DevlinCLT19}, we have three additional types of special tokens used as the inputs to the PLM:
\begin{itemize}
    \item $[$PRO$]$ (Prompt): the placeholder for continuous prompt embeddings;
    \item $[$TMSK$]$ (Token Mask): the token mask of the input texts for optimizing the auxiliary MLM loss;
    \item $[$OMSK$]$ (Output Mask): the mask as a placeholder to generate the output result.
\end{itemize}
For a better understanding, please refer to an example for single-sentence classification in Figure~\ref{fig:framework}. Here, ``[TMSK]'' is only applied to a small portion of the input texts for MLM. ``[OMSK]'' is used for generating outputs, rather than the ``[CLS]'' token. 
Hence, no additional parameters are introduced to our model for prompt learning.

As the parameters w.r.t. ``[PRO]'' tokens need to be learned for a given task, the lack of training data in few-shot learning still brings some burdens. We initialize prompt embeddings to be the pre-trained representations of~\emph{universal task-invariant prompts}.
Here, the universal task-invariant prompt for single-sentence  tasks is ``it is''; and  ``?'' for sentence-pair tasks. Note that the ``[PRO]'' and ``[OMSK]'' tokens are placed between the two pieces of texts to better capture the relations between the sentence pair. Refer to examples in Table~\ref{tab:pexample}. This setting can be viewed as the~\emph{knowledge prior} for prompt embeddings. During model training, the representations of prompts can be automatically adapted to specific tasks. In the experiments, we further show that it is unnecessary to design task-specific prompts for our approach. Despite the usage of such prompts, our method should be regarded as~\emph{task-invariant prompt-free} because no manual work is required for designing different prompts for specific tasks.


\begin{figure}
\centering
\includegraphics[width=.46\textwidth]{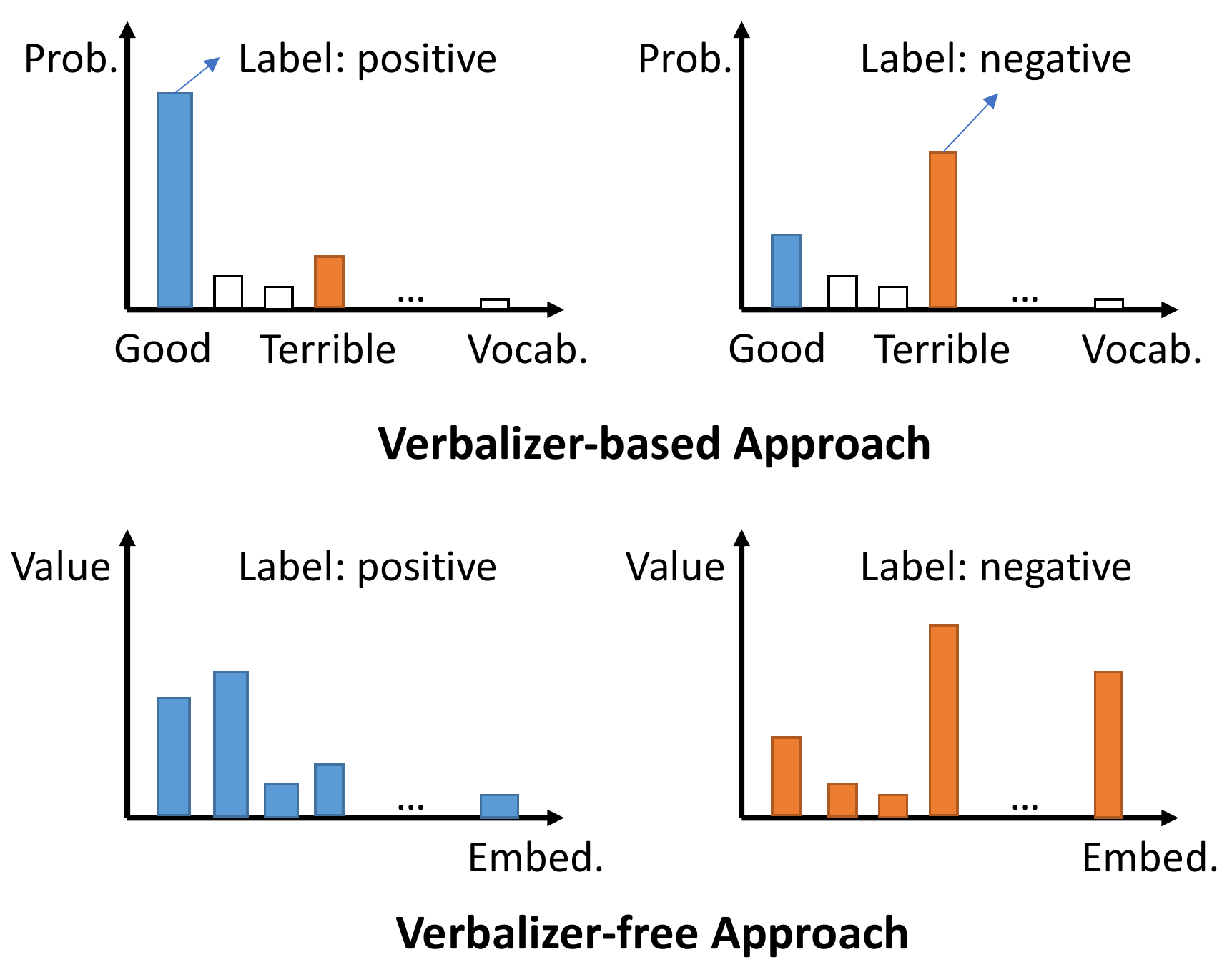}
\vspace{-.25em}
\caption{A simple comparison between~\emph{verbalizer-based} and~\emph{verbalizer-free} approaches w.r.t.the  model outputs (which are the output token probability of the masked language token and the dense ``[OMSK]'' embeddings, respectively). The underlying task is review sentiment analysis.}
\vspace{-.25em}
\label{fig:verbalizer}
\end{figure}

\subsection{Verbalizer-free Class Mapping}

A common property of existing prompt-based approaches is that they require handcrafted verbalizers to establish mappings between tokens and class labels~\cite{DBLP:conf/eacl/SchickS21,DBLP:conf/naacl/SchickS21,DBLP:journals/corr/abs-2103-10385}. We suggest that this practice might be sub-optimal. Consider the example on review analysis in Figure~\ref{fig:verbalizer}. \emph{Verbalizer-based} approaches generate the distributions over the entire vocabulary (which may contain over 10 thousand words), and only pay attention to the probabilities of very few words (such as ``good'' and ``terrible'' in our case). The semantic association between words is also ignored to a large extent. For example, the probabilities of words such as ``nice'', ``fantastic'', ``bad'' and ``horrible'' are also strong indicators of class labels.
If we replace the high-dimensional, sparse distributions with lower-dimensional, dense representations, the generalization ability and the flexibility of the underlying model can be largely increased.

In our work, we propose a novel~\emph{verbalizer-free} approach to generate model outputs based on~\emph{PCCL}.
During training, denote $\mathcal{B}$ as the collection of instances in a batch ($\mathcal{B}\subset\mathcal{D}$).
Each instance $i\in\mathcal{B}$ can be treated as an~\emph{anchor}, with the label denoted as $y_i$. We also have the~\emph{positive set} $P(i)$ and the~\emph{negative set} $N(i)$ w.r.t. the instance $i$ and the batch $\mathcal{B}$:
\begin{equation}
\begin{split}
P(i)=\{j\vert j\neq i, y_j=y_i, j\in\mathcal{B}\}\\
N(i)=\{j\vert y_j\neq y_i, j\in\mathcal{B}\}
\end{split}
\end{equation}

Let $\vec{z}_i$ be the $l_2$-normalized embedding of the ``[OMSK]'' token of the last layer of the underlying PLM (before the~\emph{softmax} function).
In the context of contrastive learning, we aim to maximize the within-class similarity $s_{i,p}=\vec{z}^T_i\cdot\vec{z}_p$ where $p\in P(i)$, and also minimize the between-class similarity $s_{i,n}=\vec{z}^T_i\cdot\vec{z}_n$ where $n\in N(i)$.
Following previous supervised contrastive learning models~\cite{DBLP:conf/nips/KhoslaTWSTIMLK20,DBLP:journals/corr/abs-2104-08821}, it is straightforward to derive the~\emph{sample-wise} contrastive loss:
\begin{equation}
\mathcal{L}_{CL}(i)=-\log\frac{\exp(s_{i,p}/\tau)}{\exp(s_{i,p}/\tau)+\exp(s_{i,n}/\tau)}
\end{equation}
where $\tau$ is the temperature value. When multiple instances in $P(i)$ and $N(i)$ are considered, we re-write $\mathcal{L}_{CL}(i)$ as follows:
\begin{equation}
\mathcal{L}_{CL}(i)=-\log\sum_{p\in P(i)}\frac{\exp(s_{i,p}/\tau)}{\sum_{a\in A(i)}\exp(s_{i,a}/\tau)}
\end{equation}
where the collection $A(i)=\mathcal{B}\setminus\{i\}$. This gives the model more generalization abilities in that multiple within-class and between-class similarity values are averaged, thus making the learned decision boundary smoother.
The partial gradient of $\mathcal{L}_{CL}(i)$ is given by:
\begin{equation}
\begin{split}
\frac{\partial\mathcal{L}_{CL}(i)}{\partial\vec{z}_i}=
& \frac{\sum_{p\in P(i)}\exp({s_{i,p}/\tau})\vec{z}_p
+\sum_{n\in N(i)}\exp({s_{i,n}/\tau})\vec{z}_n}{\tau\cdot\sum_{a\in A(i)}\exp({s_{i,a}/\tau})}\\
& -\frac{\sum_{p\in P(i)}\exp({s_{i,p}/\tau})\vec{z}_p}{\tau\cdot\sum_{p\in P(i)}\exp({s_{i,p}/\tau})}
\end{split}
\end{equation}

Minimizing $\mathcal{L}_{CL}(i)$ alone may be insufficient as it does not consider sample difficulty. For example, if $s_{i,p}=0.2$ and $s_{i,p^{'}}=0.95$ where $p,p^{'}\in P(i)$.
The model should pay more attention to $s_{i,p}$ to reach the optima, and less attention to $s_{i,p^{'}}$ to avoid model over-fitting.
Inspired by~\cite{DBLP:conf/cvpr/SunCZZZWW20}, we introduce~\emph{pair-wise relaxation factors} and propose a new loss function named~\emph{Pair-wise Cost-sensitive Contrastive Loss} (\emph{PCCL}) as follows:
\begin{equation}
\mathcal{L}_{PCCL}(i)=-\sum_{p\in P(i)}\log\frac{\exp(\alpha_{i,p}\cdot s_{i,p}/\tau_p)}{\mathcal{Z}(i)}
\end{equation}
where $\mathcal{Z}(i)$ is the normalization factor:
\begin{equation}
\mathcal{Z}(i)=\sum_{p\in P(i)}\exp(\frac{\alpha_{i,p}}{\tau_p}s_{i,p})
+\sum_{n\in N(i)}\exp(\frac{\alpha_{i,n}}{\tau_n}s_{i,n})
\end{equation}
$\alpha_{i,p}$ and $\alpha_{i,n}$ are~\emph{pair-wise relaxation factors} with the definitions formulated as follows:
\begin{equation}
\begin{split}
\alpha_{i,p}=\max\{0, 1+m-s_{i,p}\}\\
\alpha_{i,n}=\max\{0, s_{i,n}+m\}  
\end{split}
\end{equation}

\begin{figure}
\centering
\includegraphics[width=.5\textwidth]{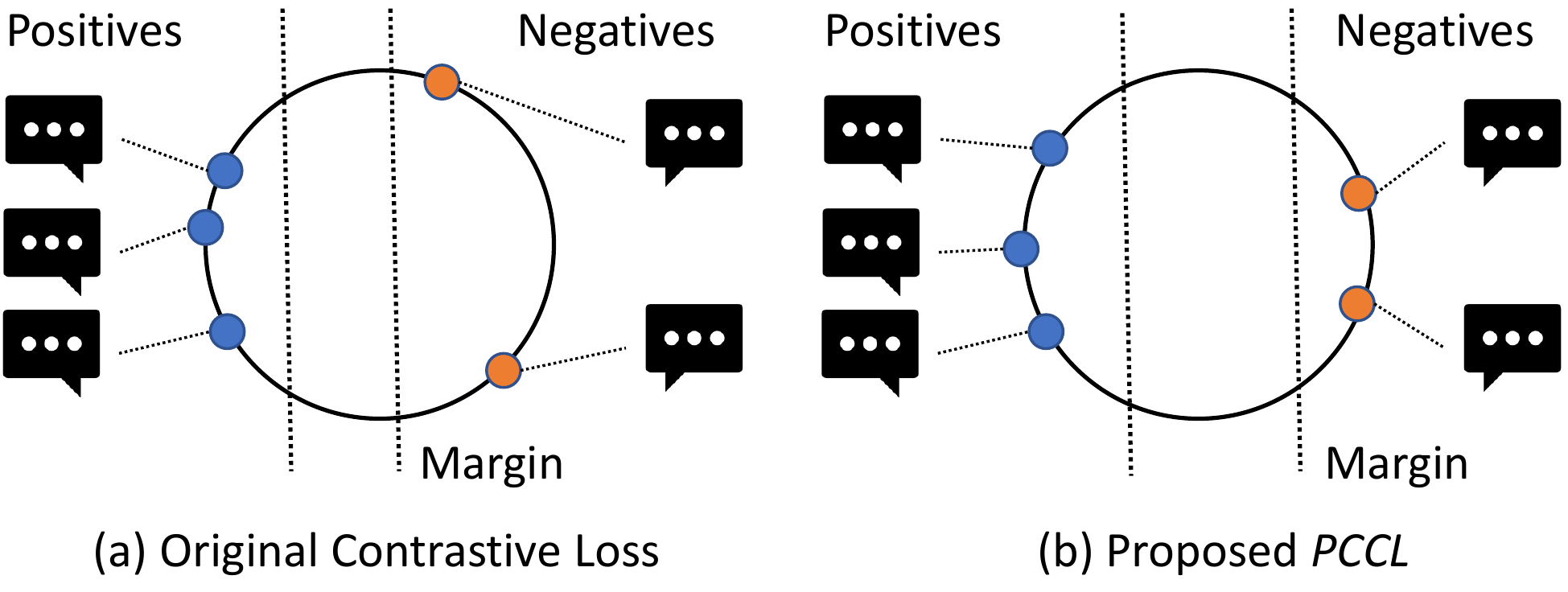}
\caption{Illustration of how~\emph{PCCL} improves the learning process of ``[OMSK]'' embeddings of the last transformer encoder layer for review sentiment analysis.
\emph{PCCL} enlarges the margins between embeddings of instances from different classes, and the margins among embeddings of different negative samples w.r.t. the anchor.}
\label{fig:example}
\end{figure}

Comparing to the original $\mathcal{L}_{CL}(i)$, two new features are added to~\emph{PCCL}.
Inside $\alpha_{i,p}$ and $\alpha_{i,n}$, a margin factor $m$ is employed to expect that $s_{i,p}>1-m$ and $s_{i,n}<m$. Hence, there is a relaxed margin between $s_{i,p}$ and $s_{i,n}$. The usage of $\alpha_{i,p}$ and $\alpha_{i,n}$ also makes the model focus on learning hard cases and avoid over-fitting on easy cases.
Another empirical setting is that we use separate temperatures $\tau_p$ and $\tau_n$ for within-class and between-class similarities, instead of a uniform temperature $\tau$. We further set $\tau_p=\xi\cdot\tau_n$ ($\xi>1$) to give more relaxations on positive samples in order to make the within-class similarities not too large, as it is easy to see:
\begin{equation}
\frac{\alpha_{i,p}}{\tau_p}s_{i,p}=\frac{\alpha_{i,p}}{\xi\cdot\tau_n}s_{i,p}=\frac{\tilde{\alpha}_{i,p}}{\tau_n}s_{i,p}
\end{equation}
where $\tilde{\alpha}_{i,p}=\max\{0, \frac{1}{\xi}(1+m-s_{i,p})\}$.
In this way, our few-shot learner will be less likely to over-fit to training instances. 
During the learning process, the $\vec{z}_i$ embeddings are optimized by computing the gradients $\frac{\partial\mathcal{L}_{PCCL}(i)}{\partial\vec{z}_i}$ and updating the PLM where:
\begin{equation}
\begin{split}
\frac{\partial\mathcal{L}_{PCCL}(i)}{\partial\vec{z}_i}=
& \frac{\sum_{p\in P(i)}\exp(\alpha_{i,p}{s_{i,p}/\tau_p})\vec{z}_p
+\sum_{n\in N(i)}\exp(\alpha_{i,n}{s_{i,n}/\tau_n})\vec{z}_n}{\mathcal{Z}_p(i)}\\
& -\frac{\sum_{p\in P(i)}\exp(\alpha_{i,p}{s_{i,p}/\tau_p})\vec{z}_p}{\tau_p\cdot\sum_{p\in P(i)}\exp(\alpha_{i,p}{s_{i,p}/\tau_p})}
\end{split}
\end{equation}
with $\mathcal{Z}_p(i)=\tau_p\sum_{p\in P(i)}\exp(\alpha_{i,p}{s_{i,p}/\tau_p})
+\tau_n\sum_{n\in N(i)}\exp(\alpha_{i,n}{s_{i,n}/\tau_n})$. We further provide an illustrative example in Figure~\ref{fig:example} and a brief theoretical analysis on~\emph{PCCL}.

\subsection{Theoretical Analysis of~\emph{PCCL}}

We theoretically show that \emph{PCCL} is an extension to various metric learning based loss functions.
As~\emph{PCCL} is directly extended from the supervised contrastive loss~\cite{DBLP:conf/nips/KhoslaTWSTIMLK20,DBLP:journals/corr/abs-2104-08821} by adding~\emph{pair-wise relaxation factors}, it is trivial to see that the supervised contrastive loss is a special case of~\emph{PCCL} with $\alpha_{i,p}=\alpha_{i,n}=1$ and $\tau_p=\tau_n$.

Next, we consider the triplet loss~\cite{DBLP:conf/cvpr/SchroffKP15}.
Assume that there are only one positive and one negative samples for each anchor. We simplify $\mathcal{L}_{PCCL}(i)$ as follows:
\begin{equation}
\begin{split}
\mathcal{L}_{PCCL}(i)^{'}& = \log(1 +\exp(\frac{\alpha_{i,p}}{\tau_p}s_{i,p}-\frac{\alpha_{i,n}}{\tau_n}s_{i,n})) \\
&= \log(1 +\exp(\frac{1}{\tau_n}(\frac{\alpha_{i,p}}{\xi}s_{i,p}-\alpha_{i,n}s_{i,n}))
\end{split}
\end{equation}

If we set a small value for $\tau_n$ (close to 0, which is the case as shown in the experiments), then the value of $\frac{1}{\tau_n}(\frac{\alpha_{i,p}}{\xi}s_{i,p}-\alpha_{i,n}s_{i,n})$ is large. We have: 
\begin{equation}
\begin{split}
& \mathcal{L}_{PCCL}(i)^{'}
\approx \frac{1}{\tau_n}(\frac{\alpha_{i,p}}{\xi}s_{i,p}-\alpha_{i,n}s_{i,n})\\
& = \frac{1}{\tau_n}(\frac{\alpha_{i,p}}{\xi}\vec{z}_i^T\vec{z}_p-\alpha_{i,n}\vec{z}_i^T\vec{z}_n)\\
&  \propto -\frac{1}{2\tau_n}(\frac{\alpha_{i,p}}{\xi}\Vert \vec{z}_i-\vec{z}_p\Vert^2-\alpha_{i,n}\Vert \vec{z}_i-\vec{z}_n\Vert^2)\\
\end{split}
\end{equation}

Approximately, the problem of minimizing $\mathcal{L}_{CCL}(i)^{'}$ is equivalent of optimizing the loss function $\mathcal{L}_{TL}(i)$ (with the margin omitted) as follows:
\begin{equation}
\mathcal{L}_{TL}(i)=\alpha_{i,n}\Vert \vec{z}_i-\vec{z}_n\Vert^2-\frac{\alpha_{i,p}}{\xi}\Vert \vec{z}_i-\vec{z}_p\Vert^2
\end{equation}
which is the triplet loss with the positive and negative pair-wise weights to be $\frac{\alpha_{i,p}}{\xi}$ and $\alpha_{i,n}$, respectively. 
Therefore, the triplet loss has a close connection to~\emph{PCCL}.

As for the N-pair loss~\cite{DBLP:conf/nips/Sohn16}, we consider the situation where there is only one positive sample and multiple negative samples for each anchor. We re-write $\mathcal{L}_{PCCL}(i)$ as:
\begin{equation}
\mathcal{L}_{PCCL}(i)^{''}=\log(1+\sum_{n\in N(i)}\exp(\frac{\alpha_{i,p}}{\tau_p}s_{i,p}-\frac{\alpha_{i,n}}{\tau_n}s_{i,n}))
\end{equation}
By setting $\frac{\alpha_{i,p}}{\tau_p}=1$ and $\frac{\alpha_{i,n}}{\tau_n}=1$, we simplify~\emph{PCCL} into the N-pair loss.
We can see that~\emph{PCCL} combines the advantages of both supervised learning and metric learning, specifically assigning different costs to easy and hard cases.

\subsection{Auxiliary Masked Language Modeling}

As the learning objective of~\emph{PCCL} is significantly different from the MLM task, minimizing $\mathcal{L}_{PCCL}(i)$ only may result in the~\emph{catastrophic forgetting} of the pre-training knowledge. Similar to~\citet{DBLP:conf/eacl/SchickS21,DBLP:conf/naacl/SchickS21}, we treat MLM as an auxiliary task during few-shot learning to improve the model performance on previously unseen instances. Denote the~\emph{sample-wise} MLM loss as $\mathcal{L}_{MLM}(i)$. The~\emph{sample-wise} overall loss function $\mathcal{L}(i)$ can be written as follows:
\begin{equation}
\mathcal{L}(i)=\lambda\cdot\mathcal{L}_{PCCL}(i)+ (1-\lambda)\cdot\mathcal{L}_{MLM}(i)
\end{equation}
where $\lambda$ is a pre-defined balancing hyper-parameter. In Figure~\ref{fig:framework}, we apply the auxiliary MLM task to ``[TMSK]'' tokens and~\emph{PCCL} to  ``[OMSK]'' tokens, separately.
This practice can be viewed as performing~\emph{task-specific continual pre-training}~\cite{DBLP:conf/aaai/SunWLFTWW20} and few-shot learning at the same time.
The overall training algorithm of~\emph{\model} is summarized in Algorithm~\ref{alg:train}.

\begin{algorithm}[t]
\caption{Training Algorithm of~\emph{\model}}
\label{alg:train}
\begin{algorithmic}[1]
\WHILE{not converge}
\STATE Sample a batch of training data $\mathcal{B}$ from $\mathcal{D}$;
\FOR {each training sample $i\in\mathcal{B}$}
\STATE Augment sample $i$ by the~\emph{task-invariant} prompt;
\ENDFOR
\STATE Feed the batch $\mathcal{B}$ (with prompts) to the PLM to generate $\vec{z}_i$ for each $i\in\mathcal{B}$;
\FOR {each training sample $i\in\mathcal{B}$}
\STATE Retrieve the positive set $P(i)$ and the negative set $N(i)$;
\ENDFOR
\STATE Compute the batch loss function $\mathcal{L}=\sum_{i\in\mathcal{B}}\mathcal{L}(i)$;
\STATE Update the model parameters $\mathcal{M}$ and the representations of prompts by minimizing the loss function $\mathcal{L}$;
\ENDWHILE
\STATE \textbf{return} the fine-tuned few-shot learner.
\end{algorithmic}
\end{algorithm}

\subsection{Model Inference}

During the model inference time, because we do not tune the ``[CLS]'' prediction head, we directly take the embedding $\vec{z}_i$ of a testing instance $i$ to generate the class label $\hat{y}_i$ by comparing $\vec{z}_i$ against the $k$-nearest neighbors in the few-shot training set. When~\emph{\model} is applied to larger training sets, for better scalability, the label $\hat{y}_i$ is predicted by:
\begin{equation}
\hat{y}_i=\text{argmax}_{c\in\mathcal{C}}\vec{z}_i^T\cdot\vec{z}_c
\end{equation}
where $\mathcal{C}$ is the collection of the class labels, and $\vec{z}_c$ is the prototype embedding of the class $c\in\mathcal{C}$ (\emph{i.e.,} the averaged embedding of all training instances with the class label as $c$).
Hence, this practice is closely in line with~\emph{prototypical networks}~\cite{DBLP:conf/nips/SnellSZ17,DBLP:journals/prl/JiCYPZ20}.

\section{Experiments}
\label{sec:exp}

In this section, we conduct extensive experiments to evaluate~\emph{\model} and compare it against state-of-the-arts. We also analyze our approach in various aspects to show its superiority.

\subsection{Evaluation Datasets}

In the experiments, we evaluate~\emph{\model} over various NLP tasks and datasets. Specifically, we focus on two NLP tasks frequently used in modern IR systems, namely sentiment analysis and sentence matching. To prove that our method can be applied to other NLP tasks as well, we also consider the tasks of natural language inference and subjectivity classification. The goals of these tasks and the corresponding datasets are listed as follows:
\begin{itemize}
    \item\textbf{Sentiment Analysis}: predicting the sentiment polarity of review comments (SST-2~\cite{DBLP:conf/emnlp/SocherPWCMNP13}, MR~\cite{DBLP:conf/kdd/HuL04} and CR~\cite{DBLP:conf/acl/PangL05});
    \item\textbf{Sentence Matching}: predicting the semantic equivalence of sentences in Web corpora and questions in online forums, respectively (MRPC~\cite{DBLP:conf/acl-iwp/DolanB05} and QQP~\footnote{\url{https://www.quora.com/q/quoradata/}});
    \item\textbf{Natural Language Inference (NLI)}: predicting the semantic relations between two sentences (QNLI~\cite{DBLP:conf/emnlp/RajpurkarZLL16} and RTE~\cite{DBLP:conf/birthday/Bar-HaimDS14});
    \item\textbf{Subjectivity Classification}: predicting whether the contents of documents are subjective or objective (SUBJ~\cite{DBLP:conf/acl/PangL04}).
    
\end{itemize}

The dataset statistics are summarized in Table~\ref{tab:dataset}.
For few-shot learning, we follow the evaluation protocols in~\citet{DBLP:conf/acl/GaoFC20} to sample few-shot training and development sets from the original full training sets.
In default, we set $K=16$ and measure the
average performance in terms of accuracy across 5 different randomly sampled training and development splits.
Hence, the performance of~\emph{\model} can be rigorously evaluated with a minimal influence of random seeds or datasets.

In addition, we consider the situation where the few-shot training sets are unbalanced w.r.t. the number of training samples for each class. The settings and results are elaborated in Section~\ref{sec:unbalanced}.

\begin{table}
\centering
\begin{tabular}{l l l l}
\hline
\bf Task & \bf Dataset & \bf \#Training & \bf \#Testing\\
\hline
\multirow{3}{*}{Sentiment Analysis} & SST-2 & 6,920 & 872\\
& MR & 8,662 & 2,000\\
& CR & 1,775 & 2,000\\
\hline
\multirow{2}{*}{Sentence Matching} & MRPC & 3,668 & 408\\
& QQP & 363,846 & 40,431\\
\hline
\multirow{2}{*}{\shortstack{Natural Language Inference}} & QNLI & 104,743 & 5,463\\
& RTE & 2,490 & 277\\
\hline
Subjectivity Classification & SUBJ & 8,000 & 2,000\\
\hline
\end{tabular}
\caption{\label{tab:dataset}
Dataset statistics. We only sample $K\times\vert\mathcal{C}\vert$ instances from the original training sets to form the few-shot training and development sets.}
\end{table}

\begin{table*}
\centering
\begin{tabular}{l | l | ccc | cc | cc | c | c}
\hline

\multirow{2}{*}{\textbf{Backbone}} & \multirow{2}{*}{\textbf{Method}} & \multicolumn{3}{c|}{\bf Sentiment Analysis} & \multicolumn{2}{c|}{\bf Sentence Matching} & \multicolumn{2}{c|}{\bf NLI} & \bf Subjectivity & \multirow{2}{*}{\textbf{Avg.}}\\
& & \textbf{SST-2} & \textbf{MR} & \textbf{CR} & \textbf{MRPC} & \textbf{QQP} & \textbf{QNLI} & \textbf{RTE} & \textbf{SUBJ} &\\
\hline\hline

\multirow{8}{*}{RoBERTa} & Standard Fine-tuning & 78.62	& 76.17 & 72.48 & 64.40 & 63.01 & 62.32	& 52.28 & 86.82 & 69.51\\
\cline{2-11}
& PET & 92.06 & 87.13 & 87.13 & 66.23 & 70.34 & 64.38	& 65.56 & 91.28 & 78.01\\
& LM-BFF (Auto T) & 90.60 & 87.57 & 90.76 & 66.72 & 65.25 & 68.87	& 65.99 & 91.61 & 78.42\\
& LM-BFF (Auto L) & 90.55 & 85.51 & 91.11 & 67.75 & 70.92 & 66.22	& 66.35 & 90.48 & 78.61\\
& LM-BFF (Auto T+L) & 91.42 & 86.84 & 90.40 & 66.81 & 61.61 & 61.89 & 66.79 & 90.72 & 77.06\\
& P-tuning & 91.42 & 87.41 & 90.90 & 71.23 & 66.77 & 63.42 & 67.15 & 89.10 & 78.43\\
& WARP & 58.80 & 55.25 & 55.55 & 65.74 & 65.80 & 52.29 & 60.07 & 65.59 & 59.89\\
\cline{2-11}
& \bf~\emph{\model} & \bf 93.35 & \bf 89.43 & \bf 91.57 & \bf 72.60 & \bf 73.56 &  \bf 69.22 & \bf 67.22 & \bf 92.27 & \bf 81.24\\

\hline\hline

\multirow{8}{*}{ALBERT} & Standard Fine-tuning & 63.98 & 64.90 & 71.50 & 56.78 & 59.32 & 53.48 & 52.14 & 80.54 & 62.83\\
\cline{2-11}
& PET & 87.11 & 81.47 & 88.32 & 57.21 & 66.16 &	55.32 & 61.85 & 83.28 & 72.59\\
& LM-BFF (Auto T) & 82.60 & 83.23 & 88.48 & \bf 64.04 & 60.28 & 59.42 
& 60.42 & 84.67 & 72.75\\
& LM-BFF (Auto L) & 86.83 & 83.02 &  89.12 & 63.43 & 59.49 & 56.86 & 57.33 & 88.08 & 73.02\\
& LM-BFF (Auto T+L) & 84.40 & 82.75 & 89.52 & 62.48 & 56.48 & 57.69 & 61.09  & 88.44 & 72.85\\
& P-tuning & 85.42 & 84.32 & 82.35 & 58.76 & 57.46 & 58.97 & 55.07 & 84.32 & 70.83\\
& WARP & 66.63 & 65.59 & 72.34 & 63.48 & 58.20 & 57.45 & 53.86 & 62.41 & 62.49\\
\cline{2-11}
& \bf~\emph{\model} & \bf 89.63 & \bf 84.68 & \bf 90.39 & 63.52 & \bf 71.05 &  \bf 62.02 & \bf 61.92 & \bf 89.02 & \bf 76.52\\
\hline
\end{tabular}
\caption{\label{tab:few-shot}Comparison between~\emph{\model} and baseline methods over the testing sets in terms of accuracy (\%).}
\end{table*}

\subsection{Experimental Settings}

To verify that~\emph{\model} is effective across different PLMs, we test the large version of two popular PLMs from Hugging Face Models\footnote{\url{https://huggingface.co/models}}, namely RoBERTa~\cite{DBLP:journals/corr/abs-1907-11692} and ALBERT~\cite{DBLP:conf/iclr/LanCGGSS20}. We consider the following methods as strong baselines:
\begin{itemize}
    \item\textbf{Fine-tuning}~\cite{DBLP:conf/naacl/DevlinCLT19}: it is the standard fine-tuning approach by utilizing the ``[CLS]'' head of the PLM.
    
    \item\textbf{PET}~\cite{DBLP:conf/eacl/SchickS21,DBLP:conf/naacl/SchickS21}~\footnote{\url{https://github.com/timoschick/pet}}: it employs manually-crafted, discrete prompt templates and verbalizers for few-shot learning.
    
    \item\textbf{LM-BFF}~\cite{DBLP:conf/acl/GaoFC20}~\footnote{\url{https://github.com/princeton-nlp/LM-BFF}}: it generates templates and label words automatically. In our work, three settings of LM-BFF are used for comparison, where ``Auto T'', ``Auto L'' and ``Auto T+L'' refer to the model with automatically generated templates, label words and both, respectively.
    
    \item\textbf{P-tuning}~\cite{DBLP:journals/corr/abs-2103-10385}~\footnote{\url{https://github.com/THUDM/P-tuning}}: it employs continuous prompt embeddings generated by light-weight neural nets and fixed verbalizers for few-shot learning.
    
    \item\textbf{WARP}~\cite{DBLP:conf/acl/HambardzumyanKM20}~\footnote{\url{https://github.com/YerevaNN/WARP}}: it leverages continuous prompts to improve the model performance in fine-tuning scenarios. Specifically, it learns task-specific word embeddings concatenated to input texts, which guide the PLM to solve the specific task.
    
\end{itemize}
Specifically, PET, LM-BFF, P-tuning and WARP are recent state-of-the-art approaches for prompt-based few-shot learning.
As the experimental settings of PET, LM-BFF, P-tuning and WARP are different, in order to conduct a rigorous comparison, we re-produce the results based on their open-source codes. Hence, it is noted that the results reported in our work are slightly different from their original papers. 

Our own~\emph{\model} algorithm is implemented in PyTorch and run with NVIDIA V100 GPUs. In default, we set $\tau_p=2$, $\tau_n=1$ (with $\xi=2$), $\lambda=0.5$, $m=0.3$ and $k=3$
, and also tune the parameters over the few-shot development sets.
The model is trained with the Adam optimizer~\cite{DBLP:journals/corr/KingmaB14}, with the learning rate and the batch size tuned around $\{1e-5, 3e-5, 5e-5\}$ and $\{4, 8, 16, 32\}$, respectively. The optimization process of the auxiliary MLM task is the same as in PET.
We also study how the change of some important hyper-parameters affect the overall performance.

\subsection{Overall Performance Comparison}
The experimental results of~\emph{\model} and all baselines on eight testing sets for few-shot learning are presented in Table~\ref{tab:few-shot}.
From the experimental results, we can draw the following conclusions:
\begin{itemize}
    \item Prompt-based methods (such as PET, LM-BFF and P-tuning) outperform standard fine-tuning by a large margin. This shows that prompts are highly useful for few-shot learning over PLMs. Based on our re-production results, LM-BFF (with different settings) and P-tuning have similar performance, while PET produces slightly lower performance. As for WARP, it does not outperforms other three prompt-based methods. The most possible cause is that it does not leverage the MLM head of the PLM for prediction in the few-shot learning setting.
    
    \item In the experiments, we employ two PLMs to evaluate the effectiveness of our approach, namely RoBERTa and ALBERT. We observe that RoBERTa outperforms ALBERT, regardless of which learning algorithm is chosen. This is expected as the language modeling abilities of RoBERTa are better than ALBERT due to the larger pre-training data and parameter size. We can also see the our approach is highly general and can be effectively applied to any BERT-style PLMs.

    \item The performance gains of~\emph{\model} over all the testing sets are consistent, compared to all the state-of-the-art methods. Overall, the average improvement is around 3\% over the two PLMs. It can be seen that even without task-specific prompts and verbalizers,~\emph{\model} is capable of producing high-accuracy models with few training instances.
    
    \item We further conduct~\emph{single-tailed, paired t-tests} to compare the accuracy scores on all tasks produced by~\emph{\model} against LM-BFF and P-tuning. Experimental results show that the improvement of~\emph{\model} is statistically significant (with the $p$-value $p<0.05$).
\end{itemize}

\subsection{Detailed Model Analysis}
We further study how~\emph{\model} improves the model performance in various aspects. Here, we treat SST-2, MR, MRPC and QQP as pilot tasks to explore our method. The underlying PLM is uniformly set to be RoBERTa-large.

\begin{table}
\centering
\begin{tabular}{l | llll}
\hline
\textbf{Method}/\textbf{Task} & \textbf{SST-2} & \textbf{MR} & \textbf{MRPC} & \textbf{QQP}\\
\hline
\bf Full Implement. & \bf 93.35 & \bf 89.43 & \bf 72.60 & \bf 73.56\\
\hline
\quad w/o. auxiliary MLM & \underline{91.35} & 86.67 & 71.96 & 72.47\\
\quad w/o. $\alpha_{i,p}$ and $\alpha_{i,n}$ & 92.50 & 88.59 & 68.28 & 69.32\\
\quad w/o. similarity avg. & 92.04 & \underline{86.37} & \underline{67.11} & \underline{69.14}\\
\hline
\end{tabular}
\caption{\label{tab:ablation}Ablation study of~\emph{\model} on four tasks in terms of accuracy (\%). ``Full Implement.'' refers to the full implementation of our method. The lowest accuracy scores over each dataset are printed underlined.}
\vspace{-.75em}
\end{table}

\subsubsection{Ablation Study}

The ablation results of~\emph{\model} are shown in Table~\ref{tab:ablation}. Here, ``w/o. auxiliary MLM'' refers to the variant of~\emph{\model} without the auxiliary MLM task; ``w/o. $\alpha_{i,p}$ and $\alpha_{i,n}$'' refers to~\emph{\model} without the pair-wise relaxation factors; and ``w/o. similarity averaging'' refers to the model setting where we only consider one positive and one negative instance for each anchor (similar to the standard triplet loss). To specify, the contrastive loss for ``w/o. $\alpha_{i,p}$ and $\alpha_{i,n}$'' is $\mathcal{L}_{CL}(i)$, while the~\emph{sample-wise} loss function for ``w/o. similarity averaging'' is:
\begin{equation}
-\log\frac{\exp(\alpha_{i,p}\cdot s_{i,p}/\tau_p)}{\exp(\alpha_{i,p}\cdot{\tau_p}/s_{i,p})
+\exp(\alpha_{i,n}\cdot{\tau_n}/s_{i,n})}
\end{equation}

From the results we can see that all three techniques contribute to the overall accuracy improvement. Specifically, the auxiliary MLM task has the most influence over SST-2, while similarity averaging contributes the most to the improvement on the remaining three datasets. It shows that all techniques proposed by this work positively contribute to the improvement.

In addition, we have a relatively surprising finding on the auxiliary MLM task.
The performance drops by a large margin when we remove the MLM task for SST-2 and MR. This is because the few-shot learning ability of PLMs is largely based on the utilization of pre-trained knowledge learned by the MLM task. In~\emph{\model}, the~\emph{PCCL} objective is significantly different from MLM, hence optimizing~\emph{PCCL} alone may lead to the catastrophic forgetting of the MLM knowledge acquired during the pre-training stage. We suggest that the auxiliary MLM task in~\emph{\model} is vital for obtaining the high performance.

\begin{figure}
\centering
	\subfigure[Varying $m$.]{
		\includegraphics[width=0.485\linewidth]{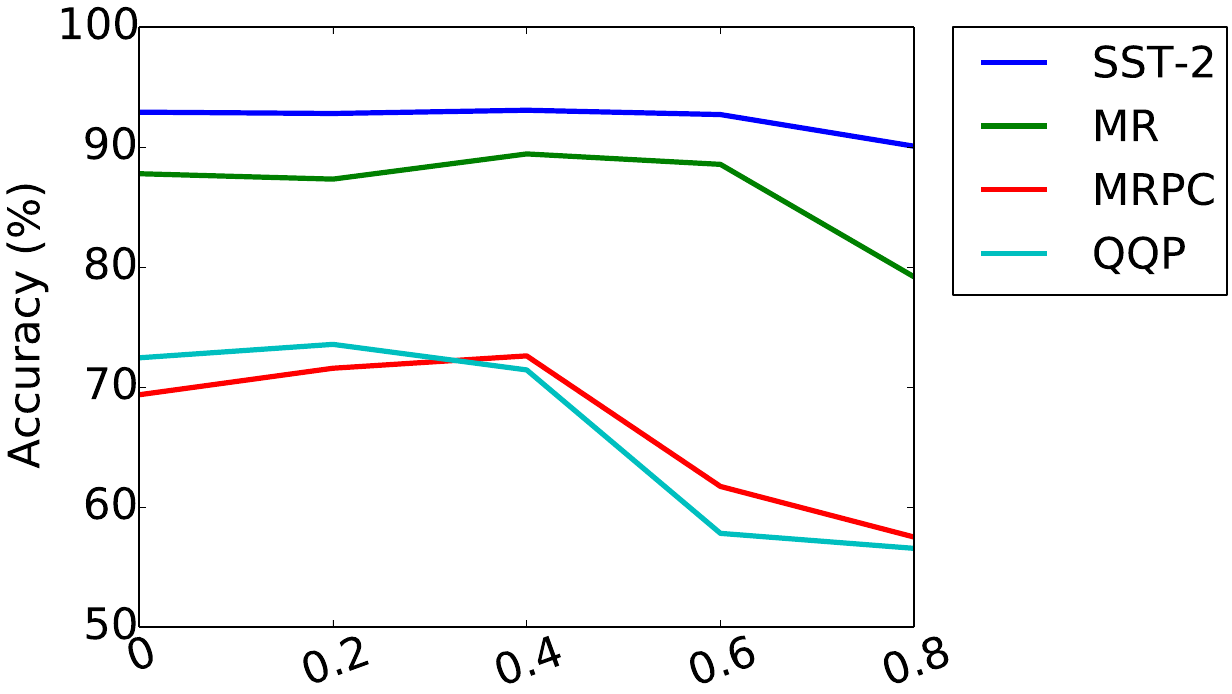}}
	\subfigure[Varying $1/\tau_n$.]{
		\includegraphics[width=0.485\linewidth]{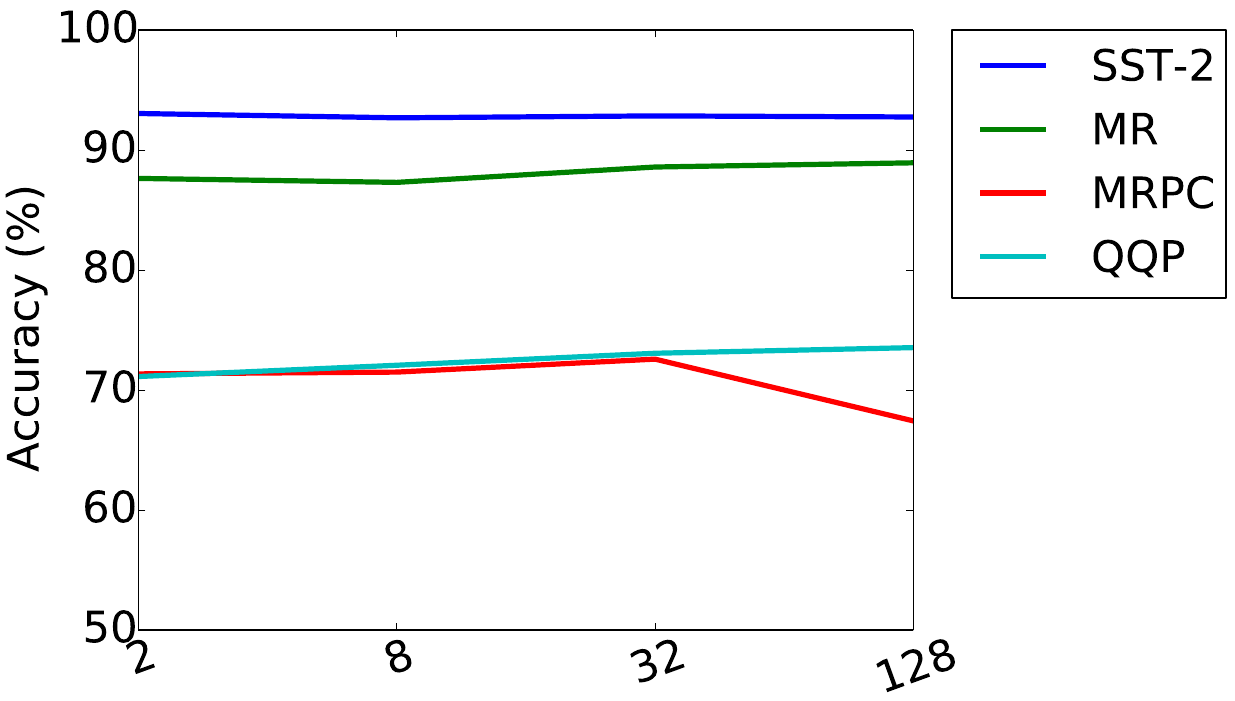}}
	\subfigure[Varying $\xi$.]{
	    \includegraphics[width=0.485\linewidth]{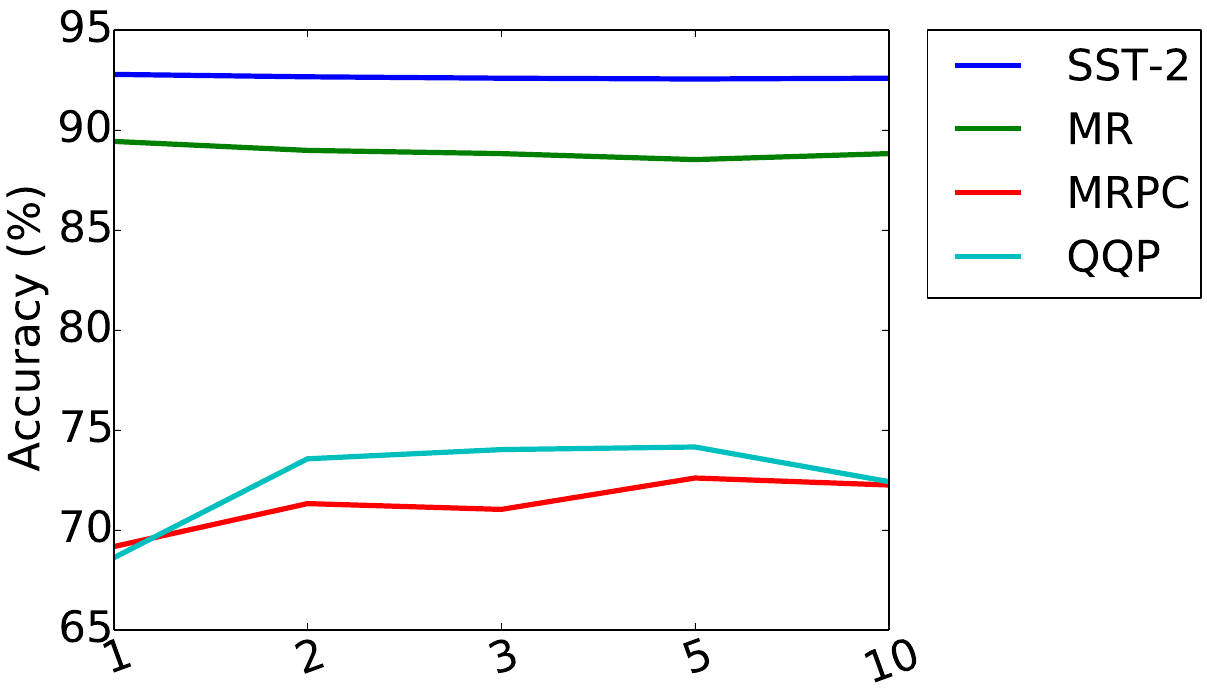}}
	 \subfigure[Varying $\lambda$.]{
	    \includegraphics[width=0.485\linewidth]{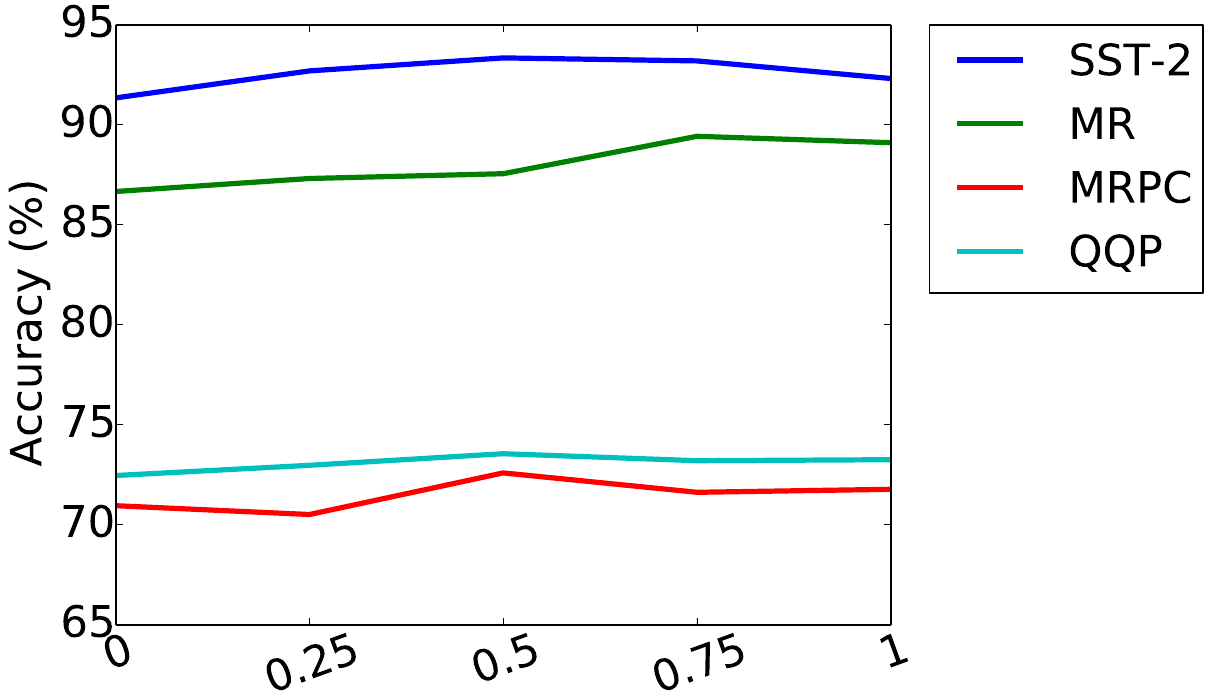}}
	\caption{Parameter analysis on hyper-parameters over four datasets in terms of accuracy.}
	\label{fig:tune}
\end{figure}

\subsubsection{Parameter Analysis}

We also show how some of the important hyper-parameters in~\emph{\model} affect the performance over the four datasets. The results are shown in Figure~\ref{fig:tune}. 
We can see the the trends are almost consistent across all the datasets. The optimal setting of the margin $m$ is around 0.2. As for the temperature, the optimal value of $\tau_n$ is around $1/8$ to $1/32$, which is different from other works where the default temperature is 1. This is probably due to the fact that we compute the total scores $\alpha_{i,p}\cdot s_{i,p}/\tau_p$ and $\alpha_{i,n}\cdot s_{i,n}/\tau_n$, which are different from those in other works in contrastive learning.
Nevertheless, the performance of~\emph{\model} is not very sensitive to the choice of the temperature, proving that~\emph{\model} is highly general for real-world applications.

We further tune the value of $\xi$. As seen in the figure, for sentence-pair tasks, the optimal $\xi$ is between 2 to 5, while easier single sentence tasks are not sensitive to this hyper-parameter.  We also try using the prototype embeddings $\vec{z}_c$ for model inference, of which the results are similar.  We suggest that when~\emph{\model} is applied to large datsets, it is suitable to predict the class label $\hat{y}_i$ by $\text{argmax}_{c\in\mathcal{C}}\vec{z}_i^T\cdot\vec{z}_c$ for better scalability.
In PET~\cite{DBLP:conf/eacl/SchickS21,DBLP:conf/naacl/SchickS21}, the auxiliary MLM task is applied with $\lambda=1e{-4}$, which is sufficiently small. In contrast to their work, we suggest that the optimal value of $\lambda$ is in the range between 0.5 to 0.75. 

\begin{table}
\centering
\begin{tabular}{l | llll}
\hline
\textbf{Batch Size}/\textbf{Task} & \textbf{SST-2} & \textbf{MR} & \textbf{MRPC} & \textbf{QQP}\\
\hline
4 & 92.80 & \bf 89.43 & \bf 72.60 & 71.84\\
8 & 92.75 & 87.98 & 71.42 & 72.92\\
16 & \bf 93.35 & 88.50 & 72.20 & \bf 73.56\\
32 & 93.28 & 89.32 & 72.42 & 73.18\\
\hline
\end{tabular}
\caption{\label{tab:batcg}Analysis of the batch size. The results show that our approach is highly effective even with a small batch size.}
\vspace{-1em}
\end{table}

\subsubsection{Analysis of Batch Size for Contrastive Learning}

Previous studies have shown that the contrastive learning process usually require a large batch size for effective representation learning~\cite{DBLP:journals/corr/abs-2011-0036}. For few-shot prompt-based learning, as we use large-scale PLMs as model backbones, the GPU memory consumption can be a significant challenge when the batch size is large. Here, we show the performance of our approach when the batch size varies, illustrated in Table~\ref{tab:batcg}.
The results show that our approach does not need a large batch size for contrastive training. We suggest that our contrastive learning setting is~\emph{fully supervised}, with strong label signals provided.
Hence, the data in a small batch can already guide the model to learn the differences between positive and negative classes.

\subsubsection{Visualizations}

\begin{figure}
  \centering
	\subfigure[Dataset: SST-2.]{
		\includegraphics[width=0.5\textwidth]{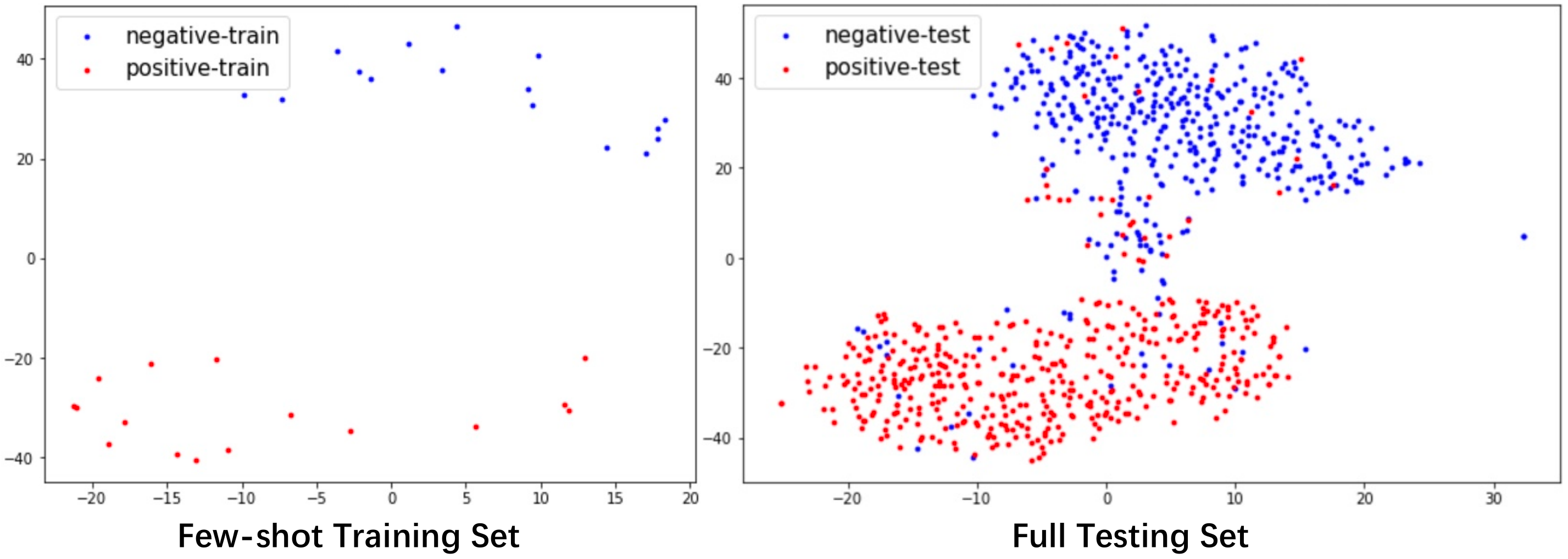}}
		
	\subfigure[Dataset: CR.]{
		\includegraphics[width=0.5\textwidth]{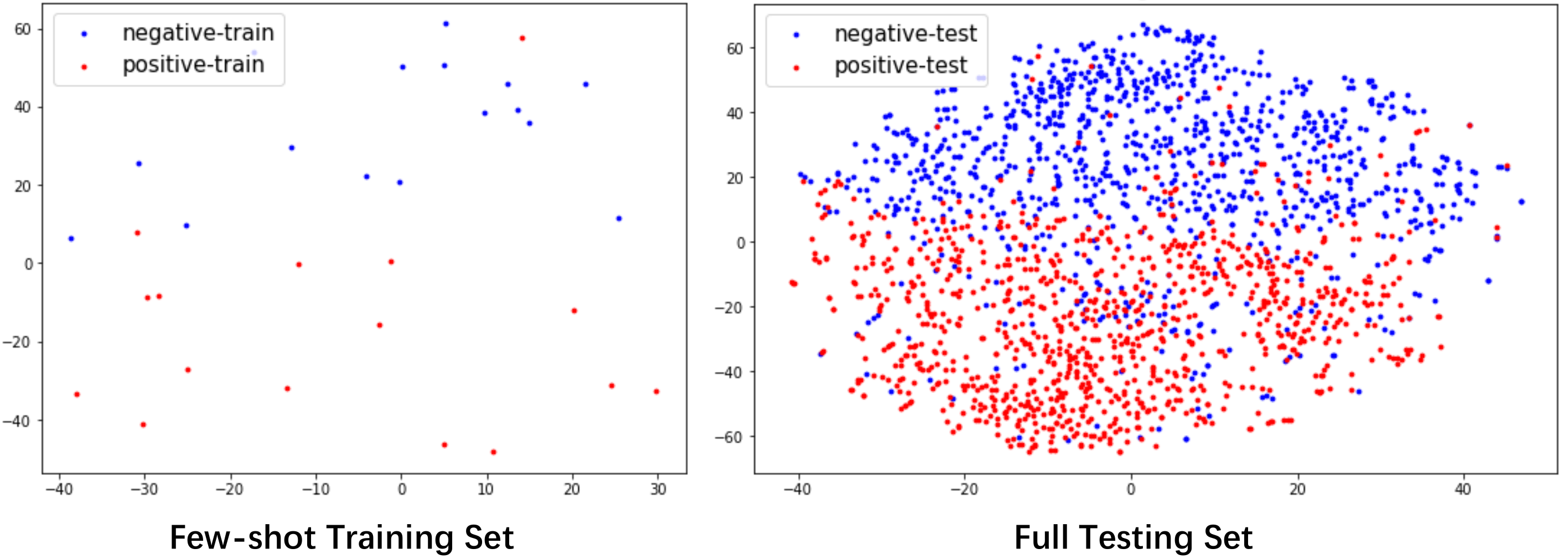}}
		
	\subfigure[Dataset: SUBJ.]{
		\includegraphics[width=0.5\textwidth]{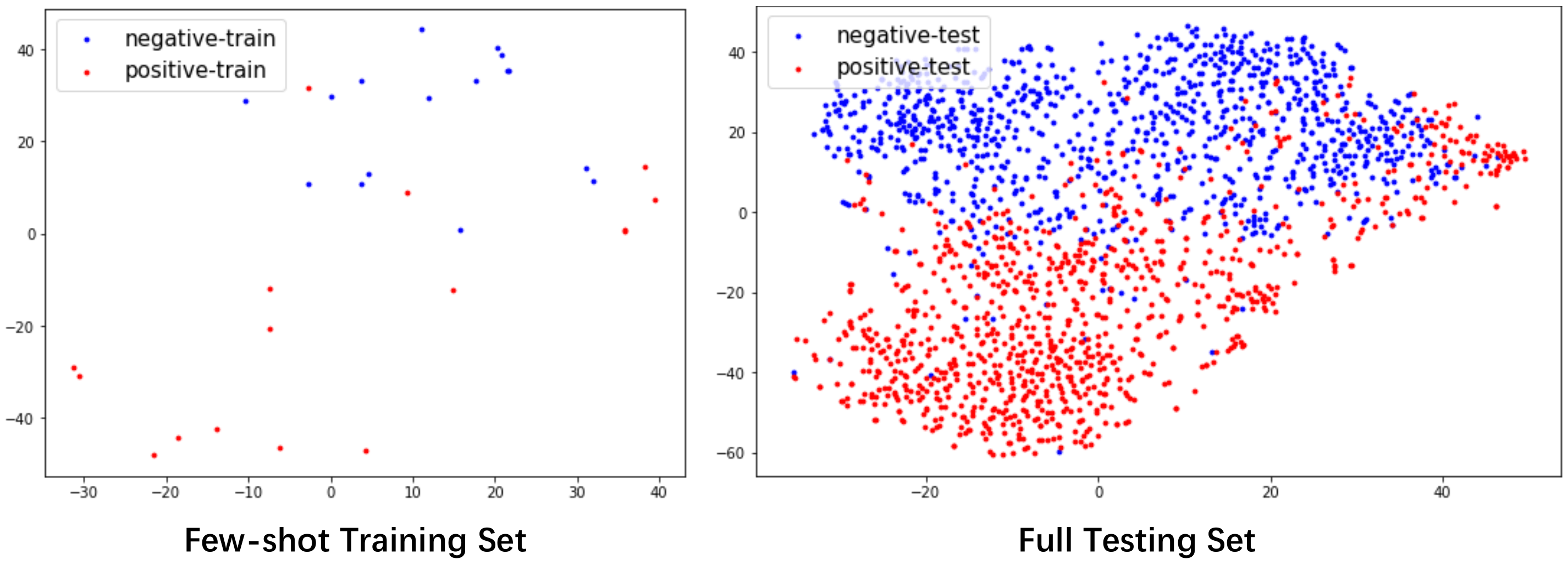}}
	
	\vspace{-.25em}	
	\caption{Visualizations of ``[OMSK]'' embeddings of the few-shot traing sets and the full testing sets of SST-2, CR and SUBJ by t-SNE. (Best viewed in color.)}
    \vspace{-.25em}	
	\label{fig:vis}
\end{figure}

To show that the generated ``[OMSK]'' embeddings are separable for text classification, we plot the embeddings of the few-shot training and testing data in SST-2, CR and SUBJ. The results are illustrated in Figure~\ref{fig:vis}. The underlying dimension reduction and visualization algorithm is t-SNE~\cite{2008Visualizing}. As seen, even reduced in two dimensions, most of the embeddings in the testing set are clearly separated, with the only $N\times K$ training samples available. Additionally, the embeddings in the few-shot training set are widely spread, showing the generalization of our algorithm. 

\subsection{Learning with Unbalanced Datasets}
\label{sec:unbalanced}

In the literature, few-shot learning is usually formulated as an~\emph{N-way-K-shot} problem. However, it may not be the case in real-world applications. In this set of experiments, we consider the situation where the few-shot training set is unbalanced w.r.t the numbers of training instances for each class. 
Following previous experiments, four binary classification tasks are used for experimental evaluation, namely SST-2, MR, MRPC and QQP. In each few-shot training set, we assume there are 8 and 24 training instances of the two classes, instead of setting $K=16$ for all the classes. The few-shot development sets are of the same size as the training sets. 

We compare~\emph{\model} against three strong baselines for few-shot learning (\emph{i.e.,} PET, LM-BFF and P-tuning). The results are shown in Table~\ref{tab:unbalanced}.
As seen,~\emph{\model} consistently outperforms these baselines by a large margin. The improvement rates are also larger than those in standard few-shot learning scenarios (as reported in Table~\ref{tab:few-shot}). This is because the contrastive learning technique in~\emph{\model} focuses on learning the distinctions between positive and negative samples, instead of tuning the MLM head only (as in previous approaches). Therefore, it is better at dealing with unbalanced few-shot learning scenarios.

\begin{table}
\centering
\begin{tabular}{l | llll}
\hline
\textbf{Method}/\textbf{Task} & \textbf{SST-2} & \textbf{MR} & \textbf{MRPC} & \textbf{QQP}\\
\hline
PET & 87.25 & 83.44 & 64.61 & 58.82\\
LM-BFF & 88.10 & 83.51 & 65.98 & 59.19\\
P-tuning & 87.92 & 83.20 & 66.64 & 61.27\\
\hline
\bf\emph{\model} & \bf 91.25 & \bf 86.52 & \bf 70.12 & \bf 65.52\\
\hline
\end{tabular}
\caption{\label{tab:unbalanced}Testing results of~\emph{\model} and baseline methods for unbalanced few-shot learning in terms of accuracy (\%).}
\vspace{-1em}
\end{table}

\begin{table}
\centering
\begin{tabular}{l | ll | ll}
\hline
\textbf{Task}/\textbf{Method} & \multicolumn{2}{c|}{\bf \emph{\model}} & \multicolumn{2}{c}{\bf {PET}}\\
\cline{2-5}
 & \bf Acc. & \bf Std. & \bf Acc. & \bf Std.\\
\hline
SST-2 & 92.91$^{*}$ & 0.56$^{*}$ & 91.28 & 1.38 \\
MR & 88.38$^{*}$  & 1.46  & 86.28 & 1.70\\
\hline
MRPC & 71.80$^{*}$ & 2.20$^{*}$ & 65.73 & 5.08 \\
QQP & 73.84$^{*}$ & 2.16$^{*}$  & 66.61 & 5.22 \\
\hline
\end{tabular}
\caption{\label{tab:pattern}Method comparison with five sets of prompts in terms of averaged accuracy (\%) and standard deviation.  $^{*}$ refers to statistical significance of higher accuracy and lower deviation at the 95\% confidence interval.}
\vspace{-.5em}
\end{table}

\subsection{Study on Task-invariance of Prompts}

In~\emph{\model}, we initialize prompt embeddings as the pre-trained representations of the universal task-invariant prompts and utilize the~\emph{verbalizer-free} mechanism to avoid the manual prompt engineering process.
In the following experiments, we aim to study whether~\emph{\model} is capable of generating more stable and accurate results using~\emph{universal task-invariant prompts}, 
compared to the non-contrastive baseline (\emph{i.e.,} PET~\cite{DBLP:conf/eacl/SchickS21,DBLP:conf/naacl/SchickS21}).

We consider two review sentiment analysis datasets: SST-2 and MR, as well as two paraphrase datasets: MRPC and QQP.
Five prompt settings are employed: the~\emph{universal task-invariant prompts} used in~\emph{\model} and the manually designed prompts used in PET~\cite{DBLP:conf/eacl/SchickS21,DBLP:conf/naacl/SchickS21}.  
In Table~\ref{tab:pattern}, we present the averaged accuracy and its standard deviation of~\emph{\model} and PET, under five different prompt settings. We can see that compared to PET,~\emph{\model} has a higher accuracy and a lower deviation when the prompts change. Hence, our task-invariant prompts are highly effective. 

This finding is different from previous works, showing that~\emph{\model} 
is not sensitive to different prompts. Hence, we suggest learning with task-invariant prompts and no verbalizers are a desirable setting that reduces the amount of human labor. 
Additionally, during the learning process, prompt embeddings can be automatically adapted to fit specific tasks.

\section{Related Work}

\label{sec:related}

In this section, we summarize related work on PLMs, prompting PLMs for few-shot learning and contrastive learning.
We also discuss how our work improves previous works from various aspects.

\subsection{Deep Contrastive Learning}
Contrastive learning~\cite{DBLP:journals/corr/abs-2011-0036} aims to learn an embedding space in which similar instances have similar embeddings while dissimilar instances fall apart. 
Contrastive learning can be either supervised or unsupervised, and achieves good performance on computer vision tasks.
In the literature, several contrastive learning objectives have been proposed, such as the triplet loss~\cite{DBLP:conf/cvpr/SchroffKP15}, the N-pair loss~\cite{DBLP:conf/nips/Sohn16}, InfoNCE~\cite{DBLP:journals/corr/abs-1807-03748} and the supervised contrastive loss~\cite{DBLP:conf/nips/KhoslaTWSTIMLK20}.
Due to its effectiveness, contrastive learning has been applied to various NLP tasks,~\emph{e.g.,} sentence representation~\cite{DBLP:journals/corr/abs-2104-08821,DBLP:conf/acl/KimYL20}, text summarization~\cite{DBLP:conf/acl/WangWXYGCW19}, aspect detection~\cite{DBLP:conf/aaai/ShiLWR21}, machine translation~\cite{DBLP:conf/acl/YangCLS19}, commonsense reasoning~\cite{DBLP:conf/acl/KleinN20}. To our knowledge,~\emph{\model}  is the first to apply contrastive learning to prompt-based few-shot learning, to make the instances of different classes more separable.

\subsection{Pre-trained Language Models}
With the two-stage pre-training and fine-tuning paradigm, 
PLMs have achieved significant improvements on various NLP tasks, frequently applied in IR systems. Readers can refer to the survey for details~\cite{DBLP:journals/corr/abs-2003-08271}.
Among these PLMs, ELMo~\cite{DBLP:conf/naacl/PetersNIGCLZ18} learns the contextual word representations by self-supervised pre-training using bidirectional LSTMs as encoders.
BERT~\cite{DBLP:conf/naacl/DevlinCLT19} is probably the most popular model, which learns the contextual representations of tokens by layers of transformer encoders.
Other PLMs based on the transformer encoder architecture include ALBERT~\cite{DBLP:conf/iclr/LanCGGSS20}, Transformer-XL~\cite{DBLP:conf/acl/DaiYYCLS19}, XLNet~\cite{DBLP:conf/nips/YangDYCSL19}, StructBERT~\cite{DBLP:conf/iclr/0225BYWXBPS20}, Big Bird~\cite{DBLP:conf/nips/ZaheerGDAAOPRWY20} and many others.
Apart from the encoder-based PLMs, the encoder-decoder architecture is used in T5~\cite{DBLP:journals/jmlr/RaffelSRLNMZLL20} and other PLMs for text generation.
The GPT model series~\cite{DBLP:conf/nips/BrownMRSKDNSSAA20} employs the auto-regressive decoder architecture for zero-shot text generation.
Our framework is highly general w.r.t. PLMs because it can be applied to any BERT-style PLMs with high accuracy.
As the neural architectures are not our major focus, we do not elaborate.

\subsection{Prompting PLMs for Few-shot Learning}
With the prevalence of GPT-3~\cite{DBLP:conf/nips/BrownMRSKDNSSAA20}, prompting PLMs for few-shot learning has become a new, popular learning paradigm. A recent survey can be found in~\citet{DBLP:journals/corr/abs-2107-13586}.
To name a few, PET~\cite{DBLP:conf/eacl/SchickS21,DBLP:conf/naacl/SchickS21} turns text classification into cloze-style problems and use manually-defined prompts to provide additional task guidance. To facilitate automatic prompt discovery,~\citet{DBLP:conf/acl/GaoFC20} generate prompts and label words from the T5 model~\cite{DBLP:journals/jmlr/RaffelSRLNMZLL20}. In addition,~\citet{DBLP:journals/tacl/JiangXAN20} also mine high-performing prompts from the training corpus.
AutoPrompt~\cite{DBLP:conf/emnlp/ShinRLWS20} employs gradient searching to detect prompts from the text corpus. However, these approaches focus on discrete prompts only and the detected prompts may not be human-understandable.

For continuous prompts, P-tuning~\cite{DBLP:journals/corr/abs-2103-10385} learns continuous prompt embeddings with differentiable parameters for GPT-based models.
The update version, P-tuning v2~\cite{DBLP:journals/corr/abs-2110-07602} extends P-tuning to different scales of PLMs and NLP tasks.
Prefix-tuning~\cite{DBLP:conf/acl/LiL20} extends the usage of continuous prompts for text generation tasks. \citet{DBLP:journals/corr/abs-2108-04106} propose a noisy channel model for prompt learning over multiple prompts.
WARP~\cite{DBLP:conf/acl/HambardzumyanKM20} leverages continuous prompts to improve the model performance in fine-tuning scenarios. Knowledgeable prompt-tuning~\cite{DBLP:journals/corr/abs-2108-02035} optimizes the verbalizer construction process by integrating the knowledge from knowledge bases.
Our work further applies contrastive learning to making the few-shot learner fully verbalizer-free, without defining task-specific prompts.

\section{Conclusion and Future Work}
\label{sec:con}

In this work, we present an end-to-end~\emph{\fullmodel} (\emph{\model}) framework that enables few-shot learning for PLMs without designing any task-specific prompts and verbalizers.  In~\emph{\model}, we employ task-invariant continuous prompt encoding and the~\emph{Pair-wise Cost-sensitive Contrastive Loss} (\emph{PCCL}) to train the model. 
Specifically, task-invariant prompt encoding eases the process of hand-crafting prompts, while~\emph{PCCL} learns to distinguish different classes and makes the decision boundary smoother by assigning different costs to easy and hard cases.
We also give a theoretical analysis on~\emph{PCCL}.
Experiments over eight public datasets show that~\emph{\model} consistently outperforms state-of-the-art methods. 

Future work of~\emph{\model} includes: i) extending the~\emph{\model} framework to other tasks such as named entity recognition, machine reading comprehension, text ranking and text generation; ii) combining~\emph{\model} with transfer learning to improve the model performance in low-resource scenarios.

\end{document}